\documentclass[10pt,twocolumn,letterpaper]{article}

\usepackage{cvpr}
\usepackage{times}
\usepackage{epsfig}
\usepackage{graphicx}
\usepackage{amsmath}
\usepackage{amssymb}
\usepackage{multirow}
\usepackage{subfigure}
\usepackage[table]{xcolor}
\usepackage{gensymb}
\usepackage{booktabs}

\newcommand{\url}{}
\definecolor{Gray}{gray}{0.9}
\cvprfinalcopy 
\newcommand{\tbf}[1]{\textbf{#1}}
\def\mbf#1{\mathbf{#1}}
\def\tbf#1{\textbf{#1}}

\newcommand{\minisection}[1]{\vspace{0.07in} \noindent {\bf #1:} }

\def\cvprPaperID{31}

\ifcvprfinal\fi
\begin{document}

\title{Pooling Faces: Template based Face Recognition with Pooled Face Images}

\author{
\begin{tabular}{c@{\extracolsep{0.6cm}}c@{\extracolsep{0.6cm}}c@{\extracolsep{0.6cm}}c@{\extracolsep{0.6cm}}c}
Tal Hassner$^{1,2}$ & Iacopo Masi$^{3}$ & Jungyeon Kim$^{3}$ & Jongmoo Choi$^{3}$ & Shai Harel$^{2}$
\end{tabular}\\
\begin{tabular}{c@{\extracolsep{0.6cm}}c}
Prem Natarajan$^{1}$ & G\'{e}rard Medioni$^{3}$\\
\end{tabular}\\
{\small $^{1}$ Information Sciences Institute, USC, CA, USA}\\
{\small $^{2}$ The Open University of Israel, Israel}\\
{\small $^{3}$ Institute for Robotics and Intelligent Systems, USC, CA, USA}\\ \vspace{-8mm}
}

\maketitle

\begin{abstract}
We propose a novel approach to template based face recognition. Our dual goal is to both increase recognition accuracy and reduce the computational and storage costs of template matching. To do this, we leverage on an approach which was proven effective in many other domains, but, to our knowledge, never fully explored for face images: average pooling of face photos. We show how (and why!) the space of a template's images can be partitioned and then pooled based on image quality and head pose and the effect this has on accuracy and template size. We perform extensive tests on the IJB-A and Janus CS2 template based face identification and verification benchmarks. These show that not only does our approach outperform published state of the art despite requiring far fewer cross template comparisons, but also, surprisingly, that image pooling performs on par with deep feature pooling.
\end{abstract}

\section{Introduction}\label{sec:intro}
Template based face recognition problems assume that both probe and gallery items are potentially represented using {\em multiple} visual items rather than just one. Unlike the term {\em set based} face recognition, {\em template} was adopted by the recent Janus benchmarks~\cite{klare2015pushing} to emphasize that templates may have heterogeneous content (e.g., images, videos) contrary to older benchmarks such as the YouTube Faces (YTF)~\cite{wolf2011ytf} in which sets contained images of a single nature (e.g., video frames). The template setting was designed to reflect many real-world biometric scenarios, where capturing a subject's facial appearance is possible more than once and using different acquisition methods. 

Ostensibly, having many images instead of one provides more appearance information which in turn should lead to more accurate recognition. In reality, however, this is not always the case. The real-world images populating these templates vary greatly in quality, pose, expression and more. Matching across templates requires that all these issues are taken under consideration to avoid skewing matching scores based on these and other confounding factors. Doing this well requires knowing which images should be compared and how to weigh the similarities of different cross-template image pairs. Beyond these, however, are also questions of complexity: How should two templates be efficiently compared without compromising (or even gaining) accuracy?

Previous work on this problem focused on the set based setting, often with the YTF benchmark, and proposed various set representations and set-to-set similarity measures. These prescribe representing face sets as anything from linear subspaces (e.g.,~\cite{hamm2008grassmann,huang2015projection}) to non-linear manifolds~\cite{chen2013improved,lu2015multi}. More recent template based methods, however, seem to prefer explicitly storing all face images over using more specialized set representations~\cite{AbdAlmageed2016multipose,chen2015unconstrained,masi2016pams,masi2016we,Swami:UMD}. Set similarity is then computed by measuring the similarities between all cross template image pairs and aggregating them into a single, template based similarity score.

We propose simple image averages (a.k.a., average pooled faces, a.k.a., 1st order set statistics) as template representations. Pooling images using pixel-wise average or median is long since known to be an effective means of correcting images, removing noise and overcoming incidental occlusions (e.g., the seminal work of~\cite{irani1995mosaic,irani1991improving,irani1993motion}). Very recently, {\em feature} pooling (rather than pooling image intensities) was proposed as an extremely useful approach for endowing existing features with invariant properties. Two such examples are scale invariance by multi-scale pooling of SIFT features~\cite{SIFT} in~\cite{dong2015domain} and pose (viewpoint) invariance by cross-pose pooling of deep features~\cite{su2015multi}. 

Rather than feature pooling, we return to pooling images directly. As we discuss in Sec.~\ref{sec:motivation}, previous work avoided relying only on this representation for face image sets and we explain why this was so. We show that the underlying requirement of successful image based pooling methods -- image alignment -- can easily be satisfied by 3D alignment techniques such as face {\em frontalization}~\cite{hassner2015effective}. Moreover, using a number of technical novelties and careful partitioning of the images in a template, based on head pose and image quality, we show that few pooled images capture facial appearances better than the original template. That is, we provide improved template matching scores but require fewer images to represent templates.

We test performance on the Janus CS2 and IJB-A datasets, using deep feature representations to encode our pooled images. We show that both face verification and identification results outperform recent state of the art. Finally, we compare our {\em image} pooling to the increasingly popular approach of deep {\em feature} pooling. Surprisingly, our results show that pooled images perform on par with pooled features, despite the fact that image alignment and averaging is computationally cheaper than deep feature extraction.

\section{Related work}\label{sec:related}

Much of the relevant work done in the past focused on the set based settings, where probe and gallery items typically comprised of multiple frames from the same video. Possibly the simplest approach to representing and matching image sets is to store the images of each set (or features extracted from them) directly, and then measure the distance between two sets by aggregating the distances between all cross-set image pairs (e.g., {\em min-dist}~\cite{wolf2011ytf}). Other, more elaborate methods designed for this purpose can broadly be categorized as belonging to four different categories. 

Set {\em Convex} or {\em Affine hulls} were both proposed as representations of face image sets. Convex hull was used by~\cite{cevikalp2010face} and then extended to the use of Affine hull in~\cite{hu2011sparse}. These methods are most effective when many images are available in each set and these hulls are well defined.

{\em Subspace methods} represent sets using linear subspaces~\cite{basri2007approximate,basri2011approximate,hamm2008grassmann,huang2015projection,kim2007discriminative}. Though the underlying assumption that all set elements lie close to a linear subspace may seem restrictive, it provides a computationally efficient representation and a natural definition for set-to-set distances: the angles between different subspaces~\cite{basri2011approximate}. Real world photos of faces, however, rarely lie on linear subspaces. Using such subspaces to represent them risks substantial loss of information and a degradation in recognition capabilities.

\begin{figure*}[t]
\centering{
\includegraphics[width=.98\textwidth]{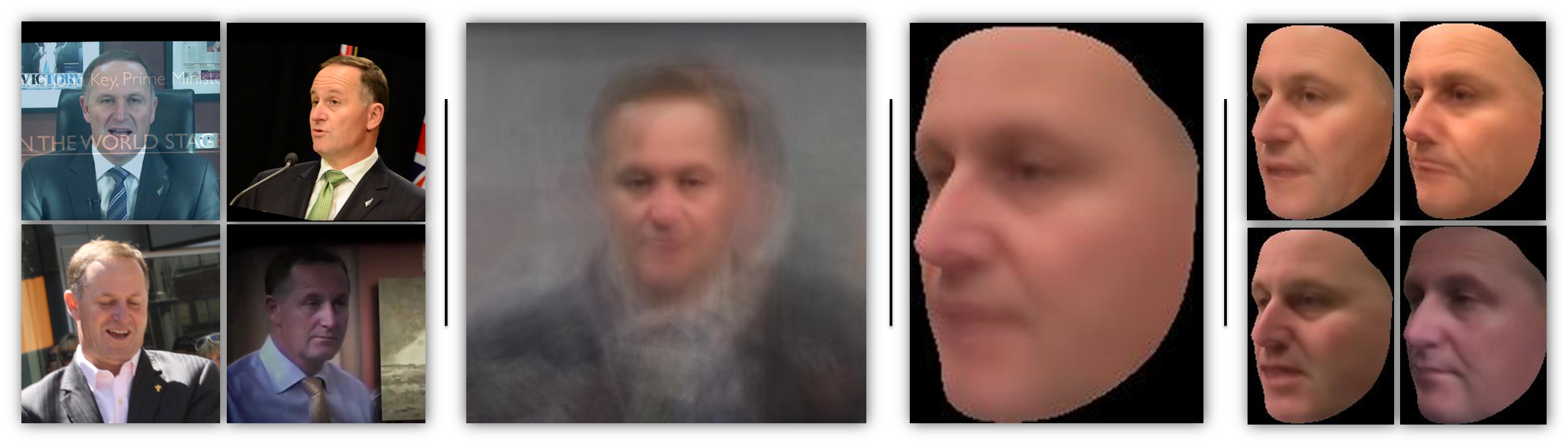}
\includegraphics[width=.98\textwidth]{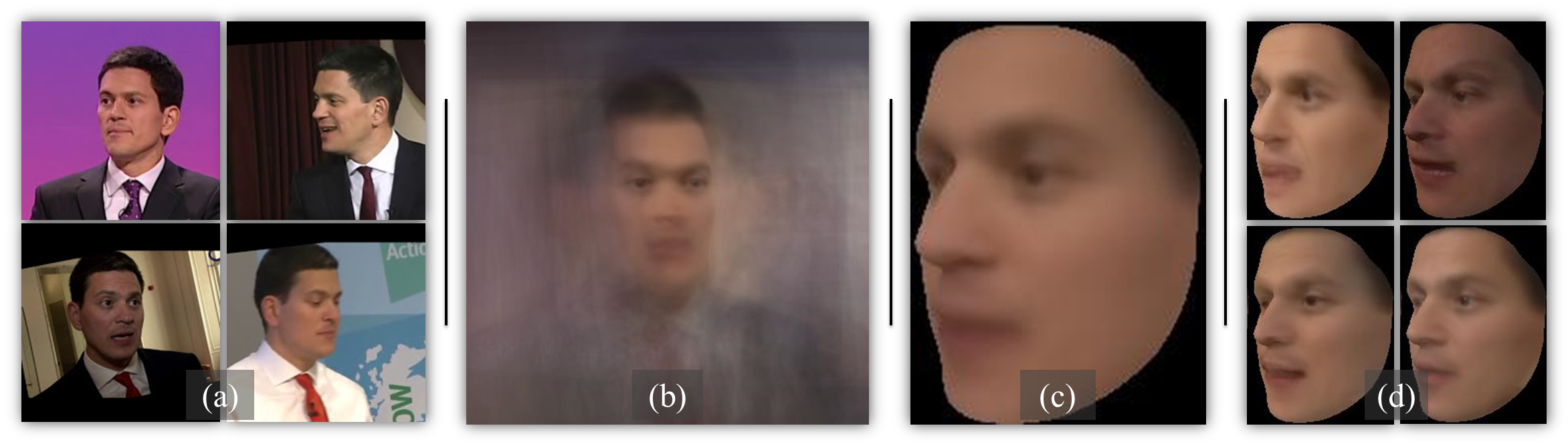}
}
\vspace{0mm}
\caption{{\bf Pooled faces}. (a) Example images from Janus~\cite{klare2015pushing} templates. (b) Averages of all in-plane aligned template images. The subjects are hardly recognizable in these averages. (c) Averages of all 3D aligned template images. Though better than (b), these are over smoothed and still hard to recognize. (d) Averages of 3D aligned images from four different face bins. These retain more high frequency information and details necessary for recognizing the subjects in the photos.}
\label{fig:quantize}
\vspace{-2mm}
\end{figure*}

When set items cannot be assumed to reside on a linear subspace, sets may still be represented by {\em non-linear manifolds}. Some examples of this approach include~\cite{chen2013improved,huang2015log,lu2015multi,wang2008manifold}. These typically require manifold learning techniques and manifold-to-manifold distance definitions which can be expensive to compute in practice. 

Finally, various {\em distribution based} representations were also considered for this purpose. Possibly the most widely used are histogram representations such as the bag of features~\cite{lazebnik2006beyond}, Fisher vectors~\cite{perronnin2007fisher} and Vector of Locally Aggregated Descriptors (VLAD)~\cite{jegou2010aggregating}. These are typically applied to sets of local descriptors, rather than images. Sets containing entire face photos were represented by 1st to n'th order statistics in~\cite{lu2013image}. Alternatively, by assuming that sets of faces are Gaussians, they were represented using covariance matrices (2nd order statistics) in~\cite{wang2012covariance} and~\cite{zhu2013point}. 

\section{Motivation: Are 1st order statistics enough?}\label{sec:motivation}
Let a (gallery or probe) face template be represented by the set of its member images (assuming that videos are represented by their individual frames), as: $\mathcal{F}=\{\mbf{I}_1,...,\mbf{I}_N \}$. where $\mbf{I}_i\in \mathbb{R}^{n\times m \times 3}$ are RGB images, aligned by cropping the bounding box centered on the face and rescaling it to the same dimensions for all images (i.e., images are assumed to be aligned for translation and scale). The 1st order statistics of this set (the average pooled face) is simply defined as:
\begin{equation}
\mbf{F} \doteq avg(\mathcal{F})=\frac{1}{N}\sum_{i=1}^N{\mbf{I}_i}\label{eq:pool}
\end{equation}

Although some of the methods surveyed in Sec.~\ref{sec:related} used 1st order statistics of face sets as part of their representations, none ventured so far as to propose using them alone, and for good reason: High order statistics and/or metric learning are required to represent and match facial appearance variations that cannot be captured effectively only by 1st order statistics. Fig.~\ref{fig:quantize} illustrates this by showing face images from a single template and their average. Apparently, averaging loses much of the information available in each individual image in favor of noise. 

Also evident in Fig.~\ref{fig:quantize} is that at least to some extent this is an alignment problem: if faces appear in exactly the same alignment (in particular, the same head pose), their average is far clearer. This was recently demonstrated in~\cite{hassner2015effective} which showed that better head pose alignments produce sharper average images. 

We go beyond the work in~\cite{hassner2015effective} and propose to cancel out variations in pose and image quality, in order to produce superior pooled faces which can be used for recognition. This, as an alternative to using high order statistics to represent face sets or expensive metric learning schemes to match them. 

Specifically, we partition a set of images into subsets containing faces which share similar appearances. We further reduce facial appearances by 3D head pose alignment. As a consequence, a face set is represented by a small collection of 1st order statistics, extracted from few subsets of the original template. Doing so has a number of attractive advantages over previous work:
\begin{itemize}
\item {\bf Reduced computational costs.} Image alignment and averaging are computationally cheaper than other existing representations. 
\item {\bf Faster matching.} Matching two templates is also quite efficient, due to the drop in the number of images representing each set. Moreover, this approach does not require expensive metric learning schemes to address appearance variations. 
\item {\bf Improved accuracy.} Despite reduced storage and computational costs, accuracy actually improves. This is likely due to the known properties of average images to reduce noise and remove incidental occlusions.
\end{itemize}

\begin{figure*}[t]
\centering{
\includegraphics[width=.98\textwidth]{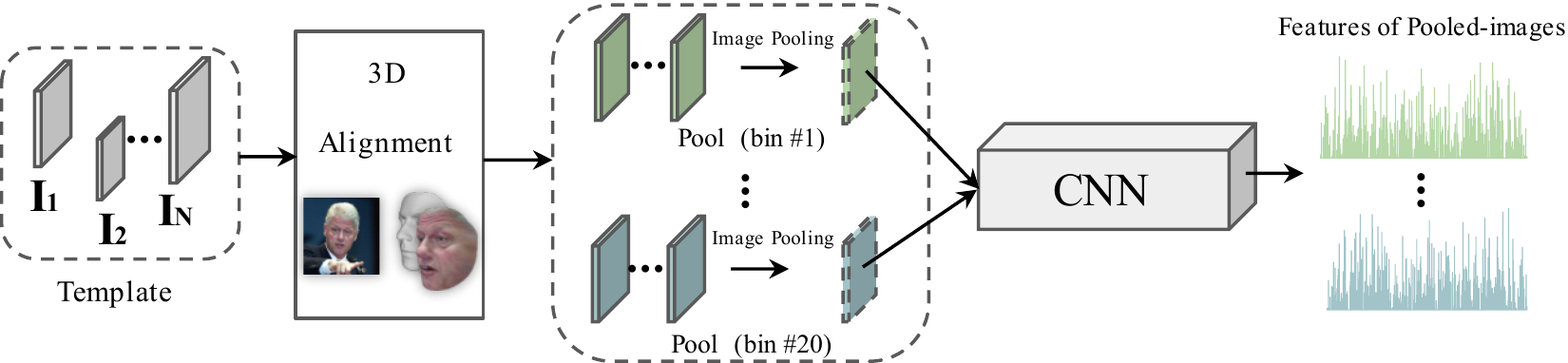}
}
\vspace{1mm}
\caption{{\bf Template representation and matching pipeline}. Illustrates the various stages of our approach. Please see text for details.}
\label{fig:pipeline}
\vspace{-2mm}
\end{figure*}

\section{Face pooling}\label{sec:method}

Our pipeline is illustrated in Fig.~\ref{fig:pipeline}. Given a face template $\mathcal{F}$, we align its images in 3D and then bin the aligned images according to pose and image quality. Images falling into the same bin are pooled, Eq.~(\ref{eq:pool}), and the pooled images are encoded using a convolutional neural network (CNN). Finally, we use these CNN features to match templates. We next describe these steps in detail. 

\subsection{Binning by head pose}\label{sec:pose}

\minisection{3D head pose estimation} The recent work of~\cite{hassner2013viewing} showed that the 6dof pose of a head appearing in a 2D image can be estimated by minimizing the geometrical distances between extracted 2D facial landmarks and their corresponding re-projected 3D landmarks on a generic 3D face model. In this work, we perform a similar process, with slight changes. 

Given a bounding box around a face, we detect 68 landmarks using CLNF~\cite{baltruvsaitis2014continuous}. Bounding boxes were estimated using the DLIB library of~\cite{king2009dlib}. We used CLNF to detect the same landmarks in a rendered image of a generic 3D face. The correspondences between the 3D coordinates on the generic model and its rendered view are obtained using the rendering code of~\cite{hassner2013viewing}. Hence, given the detected points $\mbf{p}_{i, i=1..68} \in \mathbb{R}^{2}$ on the input photo, and their corresponding points $\hat{\mbf{p}}_{i, i=1..68} \in \mathbb{R}^{2}$ on the rendered view, we have the 3D coordinates for these points, $\hat{\mbf{P}}_{i, i=1..68} \in \mathbb{R}^{3}$, on the generic face model.

Assuming the principal point is in the image center we use the 68 correspondences $(\mbf{p}_i,\hat{\mbf{P}}_i)$ to solve for the extrinsic camera parameters with the PnP method~\cite{hartley2003multiple}. This provides us with a camera matrix $\mbf{M} = \mbf{K}~\left[ \mbf{R}~\mbf{t} \right]$ minimizing the projection errors of the 3D landmarks to the landmarks detected on the input photo. The estimated pose $\mbf{M}$ is then decomposed to provide a rotation matrix $\mbf{R} \in \mathbb{R}^{3 \times 3}$ for the yaw, pitch and roll angles of the head. 

These three angles are used in three ways: roll compensation, head pose quantization and pose cancellation. Roll compensation simply means in-plane alignment of the faces so that the line between the eyes is horizontal~\cite{hartley2003multiple}. 

\minisection{Head pose quantization} Once 2D roll is eliminated, we consider only yaw angles (in practice we found pitch variations in our datasets to be small, and so only yaw angle variations were addressed; pitch angles can presumably also be used to quantize head poses to, e.g., $\pm 15^{\circ}$ pitch angle bins). Yaw angles $(|\theta|)$ are quantized into four groups, $\{(0^{\circ} \leq | \theta| < 20^{\circ}), (20^{\circ} \leq |\theta| < 40^{\circ}), (40^{\circ} \leq |\theta| < 60^{\circ}), (60^{\circ} \leq |\theta|\}$.

\minisection{Head pose cancellation} All images in all bins are then aligned in 3D to remove any remaining pose variations. We perform a process similar to the one described in~\cite{hassner2015effective}, including soft-symmetry. Unlike~\cite{hassner2015effective}, our own rendering code produced faces over a black background (the original background of the input photos was not preserved in rendering). More importantly, we found that better overall performance is obtained by rendering the faces not to frontal pose, as in~\cite{hassner2015effective}, but to a $40\degree$ (half-profile) view (Fig.~\ref{fig:quantize}). 

\subsection{Binning by image quality}\label{sec:quality}
Inspecting the images available in the templates of a recent collection such as Janus~\cite{klare2015pushing} reveals that their quality varies significantly from one photo to another. This may be due to motion blur, difficult viewing conditions or low quality camera gear. Regardless of the reason, one immediate consequence of this is that pooling low quality photos may degrade the average image. 

One way to address this is to eliminate poor images. We found, however, that doing so reduces overall accuracy, presumably because even poor photos carry some valuable information. As a consequence, these images are still used, but are pooled separately from high quality images.

\minisection{Estimating Facial Image Quality (FIQ)} We seek a FIQ measure which assigned a normalized image quality based score for a facial image $\mbf{I}$. Work on estimating image quality is abundant. We tested several existing methods for image quality estimation, ultimately choosing the {\em Spatial-Spectral Entropy based Quality} (SSEQ)~\cite{liu2014no}.

SSEQ is a {\em no-reference} image quality assessment model that utilizes local spatial and spectral entropy features on distorted images~\cite{liu2014no}. It uses a support vector machine (SVM) trained to classify image distortion and quality. A key advantage of SSEQ is that its final index allows assessing the quality of a distorted photo across multiple distortion categories. It additionally matches well with human subjective opinions of image quality~\cite{liu2014no}.

We use the code originally released for SSEQ by its authors~\cite{SSEQsoftware}. Normalized SSEQ scores were partitioned into five image quality bins, with bin limits determined empirically. Specifically, given face image the following threshold values are used to assign the image with a quality index: 
\begin{equation}
Q(\mbf{I}) = \begin{cases} 0, & \mbox{if } -\infty < SSEQ(\mbf{I}) < 0.45 \\ 
					1, & \mbox{if } 0.45 \leq SSEQ(\mbf{I}) < 0.55 \\ 
					2, & \mbox{if } 0.55 \leq SSEQ(\mbf{I}) < 0.65 \\ 
					3, & \mbox{if }  0.65\leq SSEQ(\mbf{I}) < 0.75 \\ 
					4, & \mbox{if } 0.75 \leq SSEQ(\mbf{I}) < \infty, \\                     
\end{cases}
\end{equation}
\noindent where $SSEQ(\mbf{I})$ is the FIQ measure of the input image $\mbf{I}$. 

\begin{table}[b]
\centering
\resizebox{\linewidth}{!}{  
\begin{tabular}{ll}
\toprule
Pose &$[0\degree...20\degree), [20\degree...40\degree), [40\degree...60\degree),$\\ 
(head yaw) & $[60\degree...90\degree]$\\ \hline
Quality & $(-\infty...0.45), [0.45...0.55), [0.55...0.65), [0.65...0.75),$\\
(SSEQ~\cite{liu2014no}) & $[0.75...\infty)$\\
\bottomrule
\end{tabular}
}
\vspace{2mm}
\caption{{\bf Bin indices.} The quantized pose and image quality in our template representation. Twenty bins used altogether, though few are populated in practice.}
\label{tab:bins}
\end{table}

\subsection{Representing and comparing templates}\label{sec:represent}
\minisection{Template representation} Table~\ref{tab:bins} summarizes the bin indices used in this work. All told, 20 bins are available, though, as we later show, typical templates have far fewer bins populated by images in practice. 

Pooled images are represented using deep features. Specifically, we use the VGG-19 CNN of~\cite{chatfield2014return} to encode face images. This 19 layer network was originally trained on the large scale image recognition benchmark (ILSVRC)~\cite{russakovsky2014imagenet}. We fine tune the weights of this network twice: first on original CASIA WebFace images~\cite{yi2014learning}, with the goal of learning to recognize $10,575$ subject labels of that set.

A second fine tuning is performed again using CASIA images. This time, however, training is performed following image pooling, using the process described in Sec.~\ref{sec:method}. Since CASIA has not template definitions, we use subject labels instead: We take random subsets of images from the same subject. Each subset is then treated as a template. Subjects for which only a single image exists are used in this second fine tuning step, without pooling.

\begin{figure*}[t]
\centering
\subfigure[LFW]{
\includegraphics[width=.25\textwidth]{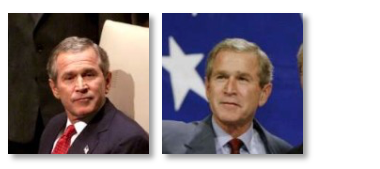}
}
\subfigure[IJB-A]{
\includegraphics[width=.72\textwidth]{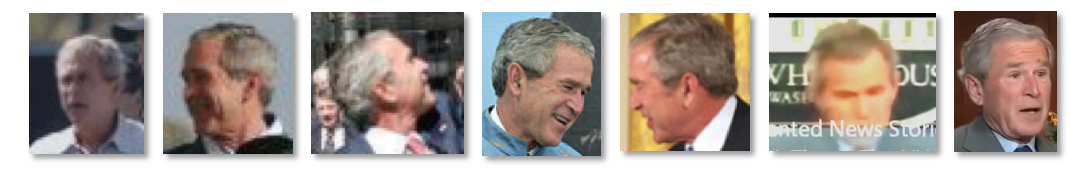}
}
\caption{Qualitative comparison between facial imagery of a subject present in both LFW and IJB-A: images in LFW has a strong bias towards media collected from the web whereas the quality of IJB-A images is far more variable. Moreover LFW benchmark considers only image-pair comparisons for face verification; while IJB-A subjects are described using image templates (sets).}
\label{fig:images}
\end{figure*}

\minisection{Matching images and templates}
A trained CNN is defined by linear functions and non-linear activations. The network is parametrized with a set of convolutional layers and fully connected layers, including values for the weights, $\mbf{W}$, and biases, $\mbf{b}$. The CNN is used to extract feature representations, $\mbf{x} = f(\mbf{I};\{\mbf{W},\mbf{b}\} )$, for each image, $\mbf{I}$ (pooled or not). We take the response produced after the fully connected layer fc7 as the image representation. Given an image $\mbf{I}_p$ in a probe template $\mathcal{P}$ and $\mbf{I}_g$ in a gallery template $\mathcal{G}$, we compute their similarity, $s(\mbf{x}^{\text{fc7}}_p, \mbf{x}^{\text{fc7}}_g)$, by taking the normalized cross correlation (NCC) of their feature vectors.

A {\em template similarity} is defined by aggregating these scores for all cross-template (pooled) image pairs (i.e., all-vs-all matching of features extracted from pooled images). We define the similarity $s(\mathcal{P},\mathcal{G})$ of two templates $\mathcal{P}$ and $\mathcal{G}$ as follows. After computing all pair-wise pooled-image level similarity scores, these values are fused using SoftMax: $s_{\beta}(\mathcal{P},\mathcal{G})$, defined in Eq.(\ref{eq:softmax}), below. The use of SoftMax here to aggregate image level similarity scores can be considered a weighted average which depends on the score to set the weights as: 

\begin{equation}
s_{\beta}(\mathcal{P}, \mathcal{G}) = \frac{\sum_{p \in \mathcal{P}, g \in \mathcal{G}}~w_{pg}~s(\mbf{x}_p, \mbf{x}_g) }{\sum_{p \in \mathcal{P}, g \in \mathcal{G}}~w_{pg} }, \quad w_{pg}\doteq \mit{e}^{\beta ~ s(\mbf{x}_p, \mbf{x}_g)}
\label{eq:softmax}
\end{equation}
As the final template similarity score, we average the SoftMax responses over multiple values of $\beta= \left [ 0 ... 20 \right ]$.

\begin{table*}[tb]
\centering
\resizebox{\textwidth}{!}{  
\begin{tabular}{|l||ccccc||ccc||cc|}
\hline
 \multicolumn{11}{|c|}{IJB-A identification (closed-set) } \\ \hline\hline
 & TPR-1\%F & TPR-0.1\%F & TPR-0.01\%F & nAUCj & FPR-85\%T & Rank-1 & Rank-5 & Rank-10 & avg-imgG & avg-imgP  \\ \hline \hline
All images &  85.9 & 71.6 & 51.3  & 63.8  & 0.7  & 82.8  & 92.1  & 94.3  & 24.3$\pm$20.8  & 7.2$\pm$13.1   \\ \hline 
Single feature pooling & 83.5  & 68.9  & 50.7  &  62.0 & 1.1  & 83.0  & 91.8  & 94.0  &    \multirow{2}{*}{{1$\pm$0}} & \multirow{2}{*}{{1$\pm$0}} \\  
Single image pooling &  61.9 &	38.4 &	19.9 & 44.3	 & 7.8 &	59.2 &	79.4 &	86.0 &  &    \\ \hline 
Random Selection per bin & 85.0  & 70.4  & 52.1  & 63.0  & 0.9  & 81.9  & 91.6  & 93.9  & \multirow{4}{*}{8.1$\pm$3.9}  &  \multirow{4}{*}{3.0$\pm$3.3} \\
Pooled features per bin &   86.1 & 72.3 & \tbf{54.1}  & 63.8  & 0.7  & 82.8  & 91.7  & 93.9  & &  \\
Pooled images per bin, wo/ft & 86.5  & 72.5  & 53.2  & 64.2  & 0.6  & 83.2  & 91.9  & 94.2   &  &  \\
Pooled images per bin, w/ft & \tbf{87.5} & \tbf{73.5}  & 53.8  & \tbf{65.0}  &  \tbf{0.5} & \tbf{84.6}  & \tbf{93.3}  & \tbf{95.1}  &  & \\\hline
\end{tabular}
}
\caption{Comparative analysis of our proposed feature pooling per bin with other baseline methods on the IJB-A identification.}
\label{tab:baselineIJBAid}
\end{table*}

\begin{table*}[tb]
\centering
\resizebox{\textwidth}{!}{  
\begin{tabular}{|l||ccccc||ccc||cc|}
\hline
 \multicolumn{11}{|c|}{JANUS CS2} \\ \hline\hline
 & TPR-1\%F & TPR-0.1\%F & TPR-0.01\%F & nAUCj & FPR-85\%T & Rank-1 & Rank-5 & Rank-10 & avg-imgG & avg-imgP  \\ \hline \hline
All images &  86.4 & 71.9 & 51.0 & 67.6 & 0.6 & 80.9 & 90.8 & 93.0   & 24.3$\pm$20.5  & 7.3$\pm$13.4   \\ \hline 
Single feature pooling & 82.9 & 68.1 & 50.9 & 64.8 & 1.3 & 79.8 & 89.8 & 91.6  &    \multirow{2}{*}{{1$\pm$0}} & \multirow{2}{*}{{1$\pm$0}} \\  
Single image pooling &  62.1 & 38.9 & 20.5 & 46.2 & 7.7 & 55.5 & 75.6 & 82.6  &  &    \\  \hline 
Random Selection per bin & 85.2 & 70.8 & 52.6 & 66.8 & 0.8 & 79.9 & 89.7 & 92.5   &  \multirow{4}{*}{8.2$\pm$3.9} & \multirow{4}{*}{3.0$\pm$3.3}  \\
Pooled features per bin &   86.5 & 73.4 & 54.0 & 67.8 & 0.6 & 81.4 & 90.5 & 92.7  &   &    \\
Pooled images per bin, wo/ft & 86.9 & 73.2 & 54.2 & 68.1 & 0.6 & 81.2 & 90.7 & 93.0   &  &  \\
Pooled images per bin, w/ft & \tbf{87.8} & \tbf{74.5}  & \tbf{54.5}  & \tbf{69.0}  &  \tbf{0.5} & \tbf{82.6}  & \tbf{91.8}  & \tbf{94.0}  &  & \\\hline
\end{tabular}
}
\caption{Comparative analysis of our proposed feature pooling per bin with other baseline methods on the JANUS CS2 splits.}
\label{tab:baselineCS2}
\vspace{-3mm}
\end{table*}

\section{Experiments}\label{sec:experiments}
We tested our pooling approach extensively on the IARPA Janus Benchmark-A (IJB-A) and JANUS CS2 benchmark. Compared to previous benchmarks (e.g., Labeled Faces in the Wild~\cite{LFWTechUpdate,LFWTech} and YTF~\cite{wolf2011ytf}) this dataset is far more challenging and diverse in its contents. In particular IJB-A brings two design novelties over these older benchmarks: 
\begin{itemize}
\item Janus faces reflect a wider range of challenges, including extreme poses and expressions, low quality and noisy images and occlusions. This is mainly due to the its design principles which emphasize heterogeneous media collections.
\item Subjects are represented by templates of images from multiple sources, rather than single images. They are moreover described by both still-images and frames from multiple videos. Throughout this paper, we therefore followed their terminology, referring to image templates rather than sets (as in the YTF collection).
\end{itemize}
Fig.~\ref{fig:images} provides a qualitative comparison between LFW and IJB-A images, highlighting the difference between a subject included in both sets: images in LFW are strongly biased towards web based production quality images, whereas IJB-A images are of poorer quality and wide pose changes. 

\subsection{Performance Metrics}
\minisection{Standard Janus verification metrics} We report the standard performance metrics for IJB-A. For both the verification and identification protocols we show different recall values (True Positive Rate) at different cut-off points in the False Positive Rate (FPR) of the ROC. The FPR is sampled at an order of magnitude less each time, ranging from TPR-1\%F (TPR at 1\% of FPR) to the most difficult point at TPR-0.01\%F (TPR at 0.01\% of FPR). 

This evaluation fits perfectly with the face verification protocol defined in IJB-A verification, as also previously done for LFW. Considering the ROC, we also show the opposite, which we believe to be more relevant in real-world scenarios: assume that we want to have a fixed recall of 85\%, the system should report what is the FPR. We denoted this metric as FPR-85T\% in the Tables. 

\minisection{Normalized Area under the Curve} We propose a novel metric which is relevant to applications where high recall is desired at very low FPR. We derive it from the ROC, as follows: we report the normalized Area under the Curve (nAUC) in a very low FPR range of $\text{FPR}=[0,...,1\%]$. We denote this metric by nAUCj in our results.

\minisection{Face identification} If the protocol allows for face identification, we also report metrics designed to assess how well each method retrieves probe subjects across a pre-defined gallery. A standard tool to measure this is the Cumulative Matching Characteristic (CMC): for IJB-A identification and JANUS CS2 we also provide the recognition rate at different ranks (Rank-1, Rank-5 and Rank-10).

\minisection{Template size} Unique to this work is its emphasis on reducing the number of images used to represent a template. As such, for each matching method we additionally report the number of pooled images used in practice (i.e., the number of populated bins in our representation, Sec.~\ref{sec:method}). We provide the average $\pm$ standard deviation (SD) for probe and gallery templates for each of the methods used. Importantly, besides reducing storage requirements, a smaller number of images results in fewer images being compared and hence faster template to template comparisons.

\subsection{Comparison with template pooling baselines}
We begin by examining the following alternative means for pooling and their effect on performance: 

\begin{itemize}

\item{\bf All images.} No pooling, all template images are used directly. To this end we use a CNN that was only fine tuned once on CASIA (was not fine tuned again using pooled images).

\item{\bf Single image pooling} Pooling all images after rendering to half-profile view into a single-averaged image per template (i.e., no image binning). 

\item{\bf Single feature pooling} Average pooling all the deep features extracted from all the images in a template into one feature vector. This follows similar techniques recently shown to be successful by, e.g.,~\cite{chen2015unconstrained,Swami:UMD}. Features were extracted using a CNN that was not trained on pooled images. 

\item{\bf Random Selection per bin} Rather than pooling the images inside a bin, we randomly select one of the images as the representative for that bin. A template therefore has the same number of images used to represent it as our own method. Here too, we used a CNN that was not fine tuned on pooled images to extract the features. 

\item{\bf Pooled features per bin} Same as single feature pooling above, but pooling features of each bin separately. 

\item {\bf Pooled images per bin (proposed)} The method described in Sec.~\ref{sec:method} and~\ref{sec:represent}. Note that for this particular approach we additionally tested the effect of fine tuning the network for each of the ten training splits in each benchmark (denoted by ``w/ft''). 

\end{itemize}

\begin{figure*}[tb]
\centering
\subfigure[]{
\includegraphics[width=.45\textwidth,clip,trim = 0mm 0mm 0mm 0mm]{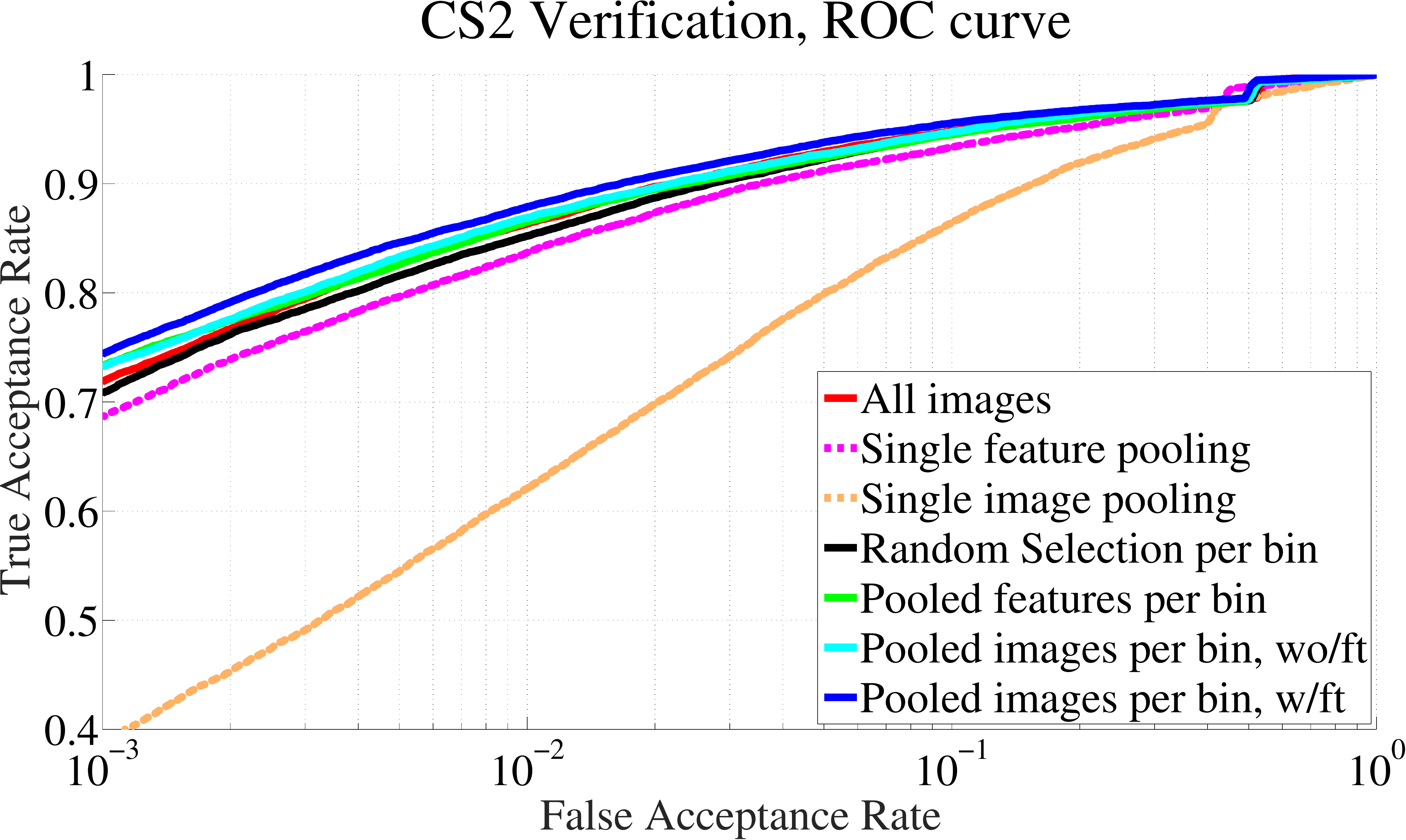}
}
\quad
\subfigure[]{
\includegraphics[width=.45\textwidth,clip,trim = 0mm 0mm 0mm 0mm]{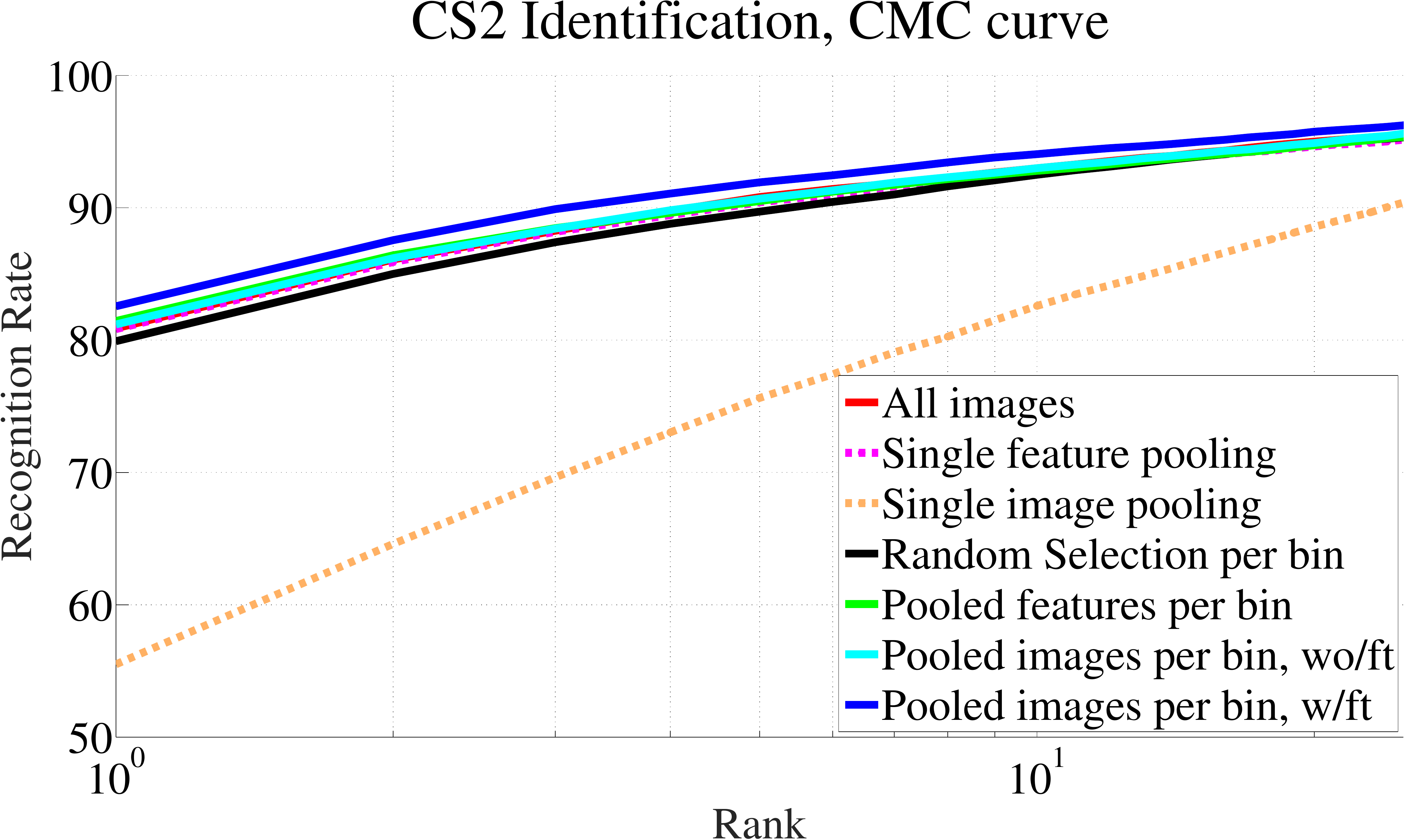}
}

\caption{(a) ROC and (b) CMC curves for Janus CS2 dataset considering all the tested pooling techniques.}
\label{fig:baselinesCS2}
\vspace{-3mm}
\end{figure*}

\begin{table*}[tb]
\centering
\resizebox{\textwidth}{!}{  
\begin{tabular}{|l||ccccc||cc|}
\hline
 \multicolumn{8}{|c|}{IJB-A verification} \\ \hline\hline
 & TPR-1\%F & TPR-0.1\%F & TPR-0.01\%F & nAUCj & FPR-85T\%  & avg-imgT$_1$ & avg-imgT${_2}$ \\ \hline \hline
All images &  80.8 & 64.0 & 33.3  & 42.6  & 1.7   & 24.3$\pm$20.8  & 7.3$\pm$13.4   \\ \hline 
Single feature pooling & 76.8  & 58.5  & \tbf{41.0}  &  40.6 & 2.8    &    \multirow{2}{*}{{1$\pm$0}} & \multirow{2}{*}{{1$\pm$0}} \\  
Single image pooling & 51.5 &	29.2 &	13.7  & 27.4 &	14.2  &  &    \\ \hline
Random Selection per bin & 79.3  & 62.2  & 33.7  & 41.8  & 2.1    &  \multirow{4}{*}{8.2$\pm$3.9} & \multirow{4}{*}{3.0$\pm$3.3}  \\
Pooled features per bin &   80.3 & \tbf{64.8} & 27.4  &  42.4  & 1.8  &   &    \\
Pooled images per bin, wo/ft & 81.0  & 62.6  & 35.2  & 42.7 & 1.7   &  &  \\
Pooled images per bin, w/ft & \tbf{81.9} & 63.1  & 30.9  & \tbf{43.1}  &  \tbf{1.4}   &  & \\\hline
\end{tabular}
}
\vspace{1mm}
\caption{Comparative analysis of our proposed feature pooling per bin with other baseline methods on the IJB-A verification.}
\label{tab:baselineIJBAverif}
\vspace{-3mm}
\end{table*}

\begin{figure*}[tb]
\centering
\subfigure[]{
\includegraphics[width=.45\textwidth,clip,trim = 0mm 0mm 0mm 0mm]{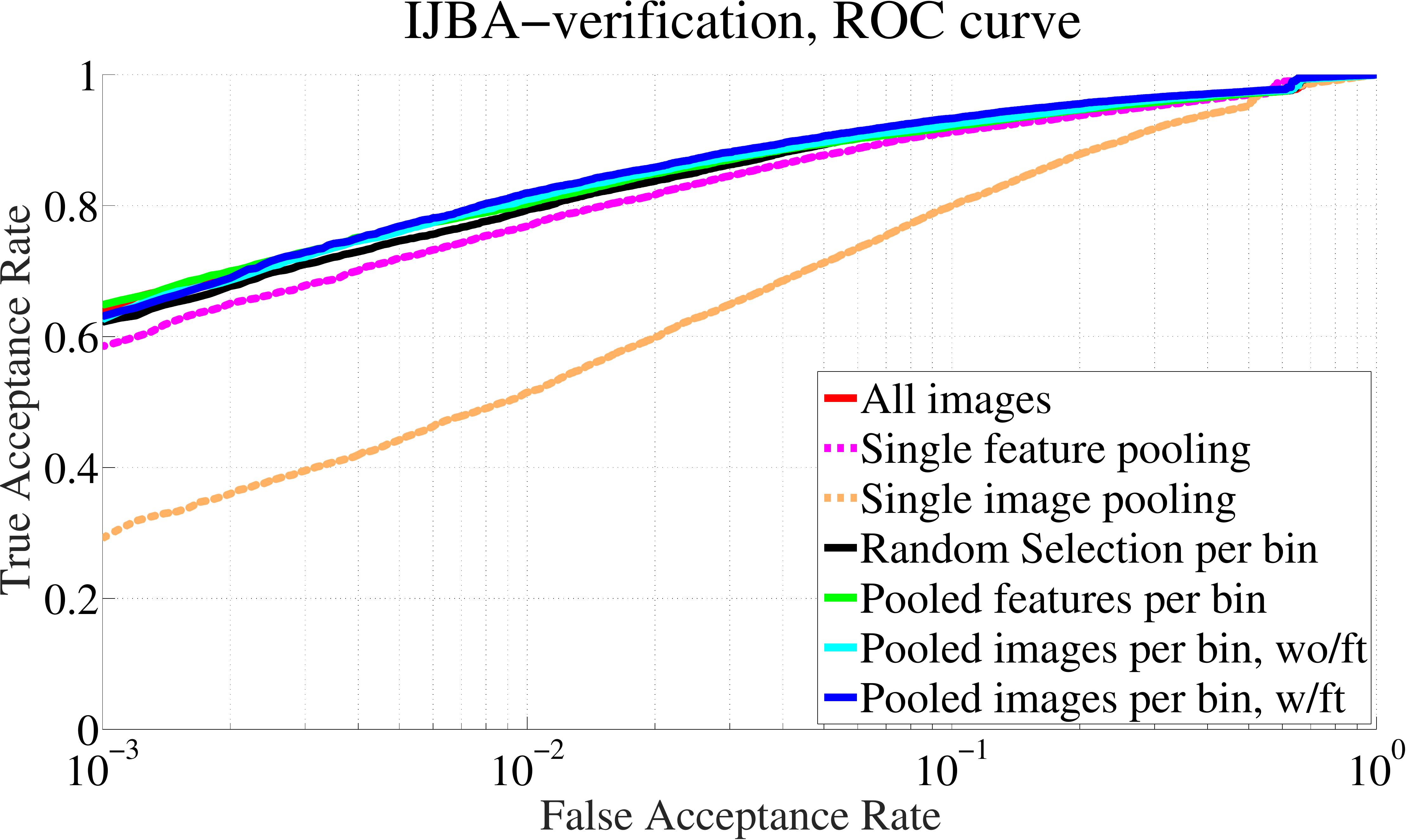}
}
\quad
\subfigure[]{
\includegraphics[width=.45\textwidth,clip,trim = 0mm 0mm 0mm 0mm]{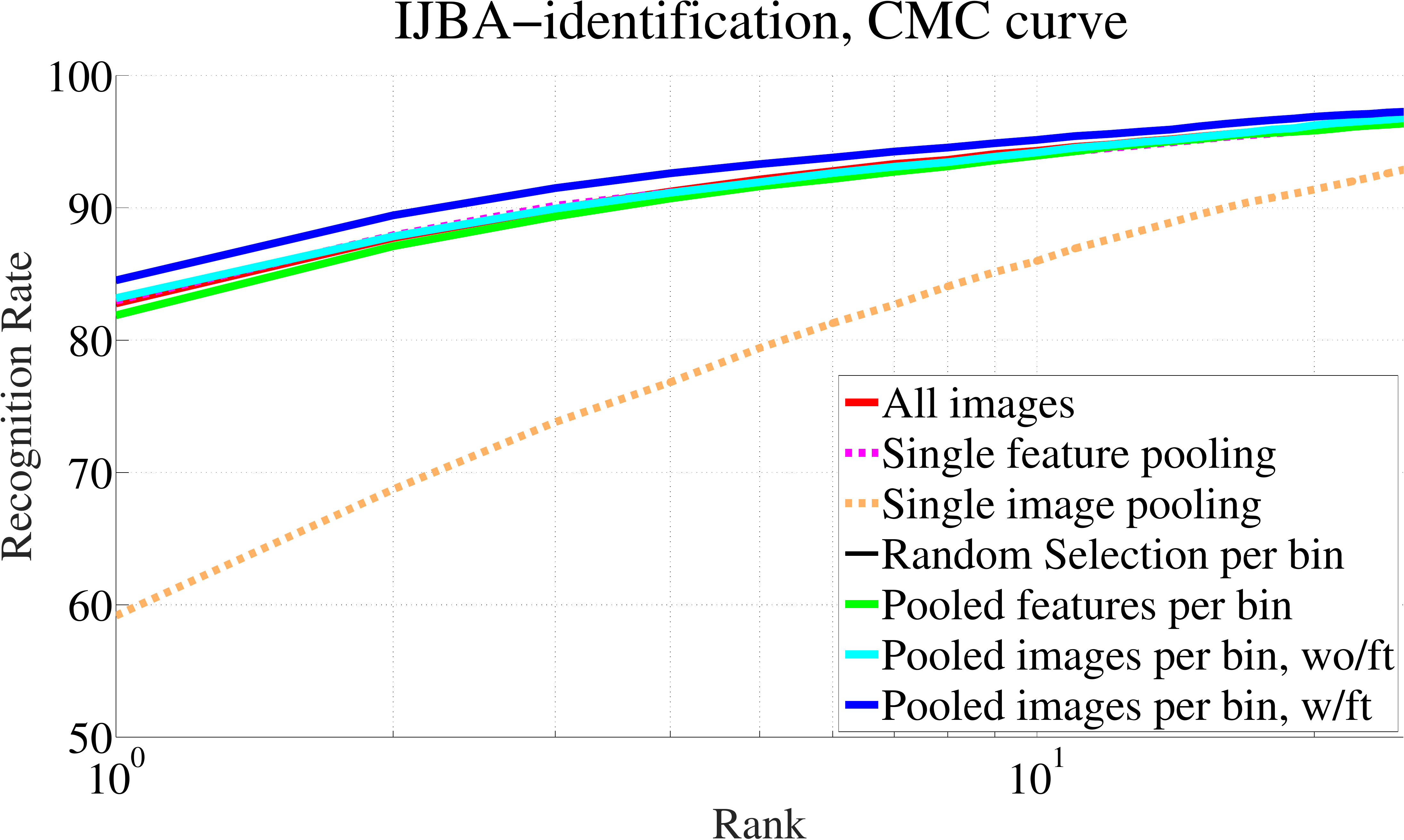}
}
\caption{(a) ROC and (b) CMC curves for IJB-A benchmark considering all the tested pooling techniques.}
\label{fig:baselinesIJBA}
\vspace{-3mm}
\end{figure*}

\subsubsection{Analysis of Performance}
All the methods presented in this paper solve the problem of comparing two sets of images and produce a score or a distance. More formally, they estimate the similarity $s(\mathcal{P},\mathcal{G})$, given a template in the probe $\mathcal{P} = \{ \mbf{x}_1, ..., \mbf{x}_P\}$ and another template in the gallery $\mathcal{G} = \{ \mbf{x}_1, ..., \mbf{x}_G\}$, where $\mbf{x}_i$ represents a feature extracted from the CNN given an image $\mbf{I}_i$. $G$ and $P$ represent the cardinality of each set. Note that if both $G=1$ and $P=1$, then the problem is reduced to the more standard LFW matching problem. What we analyze in our work is how to exploit multiple images present in a template in the framework of the Janus benchmarks. We emphasize that the Janus benchmarks do not specify fixed sizes for both $G$ and $P$, so the number of images per template varies for each testing pair, and can be as low as one.

We can interpret the pooling method as a set operator which merges together features or images, changing the cardinality of the template. For instance, considering a template in the gallery, after pooling the template, $\mathcal{G}^\prime=\text{pool}(\mathcal{G})$, its cardinality changes: $G \rightarrow  G^\prime$. As previously mentioned, this may reduce the space and time complexity required for face recognition, but it could also degrade performance if this process is not carefully performed, due to potential loss of important information.

The first baseline we examine uses no pooling, instead using all the images available in a template: this approach has complexity $\mathcal{O}(G\cdot P)$ and provides a good baseline in the Janus benchmarks, evident in Tables~\ref{tab:baselineIJBAid},~\ref{tab:baselineCS2} and~\ref{tab:baselineIJBAverif}. 

At the opposite end are methods which pool all images together into a single, compact representation (i.e., transforming template cardinality to $G^\prime=1$. This is a tremendous compression in terms of media used for recognition. One interesting observation from our analysis is that pooling all features together is much more robust than pooling all the images. This can be explained by noting that deep features are trained explicitly to recognize faces and are very discriminative. Thus averaging them does not impair matching. Moreover, such methods can evaluate a template pair at constant time as each template is represented by a single feature, yielding a complexity of $\mathcal{O}(1)$\footnote{Complexities do not reflect the time spent for performing pooling.}.

The trade-off between the two approaches is defined by compressing the media into a certain number which is more than one but still lower than the number of images in the original template, thus requiring $1 < G^\prime\ll G$. In our case $G^\prime$ corresponds to the number of populated bins, and it is always far smaller than $G$. 

Importantly, our tests show that binning and quantization play remarkable roles in partitioning the template image spaces: even naive random selection of an image within a bin improves over single feature and image pooling techniques, nearly matching the baseline performances obtained by using all template images. The complexity for these methods is equivalent to performing pair-wise bin matching, thus yielding $\mathcal{O}(G^\prime\cdot P^\prime)$.

Performance curves for all tested methods are provided in Fig.~\ref{fig:baselinesCS2} for JANUS CS2 and Fig.~\ref{fig:baselinesIJBA} for IJB-A. In particular in both figures we can see that the first order pooling methods, such as single feature/image pooling, represented by dashed curves in the graphs, are less robust compared to our proposed approach. Moreover, a big gap is clearly evident for single image pooling in the figures.

\subsection{Comparison with state-of-the-art methods}
Table~\ref{tab:sota} compares our performance to a number of existing methods on the Janus benchmarks. Our approach outperforms open-source and closed-source systems and is comparable to other published results. 

In particular we largely outperform the two baselines reported in the IJB-A dataset in~\cite{klare2015pushing}, obtained using Commercial and Government Off-The-Shelf systems (``C'' and ``G'' in Table~\ref{tab:sota}). Moreover, we consistently improve over the Open Source Biometric tool of~\cite{openBR} (ver. 0.5).

Our CNN based method improves over frontalized images encoded using the Fisher Vectors of~\cite{umd:FV} in all metrics. It further provides a better ROC for face verification in IJB-A than the recent method of~\cite{Swami:UMD}. The latter, however, achieves better results on the search protocol (identification). Interestingly, our method and~\cite{Swami:UMD} work in very different ways and provide different outcomes: we leverage template image pooling and a deeper network while~\cite{Swami:UMD} use a shallower CNN but learn a triplet similarity embedding (TSE) for each IJB-A split. This technique appears to have a better impact on identification compared to verification. Moreover, it is worth noting that~\cite{Swami:UMD} performs what we call ``single feature pooling'', which in our tests under-performed compared to the other methods we tested. 

 \begin{table}[tb]
 \centering
 \setlength\arrayrulewidth{0.1pt}
 \resizebox{\columnwidth}{!}{  
 \begin{tabular}{l||c|c|c|c|c|c||c}
 \hline
                 & C\cite{klare2015pushing} & G\cite{klare2015pushing} & \cite{openBR} & \cite{umd:FV} & \cite{80Msearch} & \cite{Swami:UMD} & Ours \\ \hline
                    \multicolumn{8}{c}{JANUS CS2} \\  \hline

TPR-1\%F        & .581 & .467 & -- & .411 &  -- &  -- & \tbf{.878}   \\
TPR-0.1\%F       & .37 & .25 &-- &  --  &  -- &  -- &  \tbf{.745}  \\
R. Rank-1         & .551 & .413  & -- & .381 &  -- &  -- &  \tbf{.826}   \\
R. Rank-5         & .694 & .571  & -- & .559 &  -- &  -- &   \tbf{.918} \\
R. Rank-10         & .741 & .624 & -- & .637 &  -- &  -- &  \tbf{.940}   \\ \hline

  \multicolumn{8}{c}{IJB-A Verification} \\ \hline

TPR-1\%F         & -- & .406 & .236 & -- & .732 & .79 & \tbf{.819}   \\
TPR-0.1\%F       & -- & .198 & .104 & -- & --   & .59 &  \tbf{.631}  \\ \hline

\multicolumn{8}{c}{IJB-A Identification} \\ \hline
R. Rank-1        & -- & .443 & .246 & -- & .820 & \tbf{.88} &  .846  \\
R. Rank-5        & --  & .595 & .595 & -- & .929 & \tbf{.95} &  .933  \\
R. Rank-10      &  -- &  --  & -- & -- &  --   & -- &  \tbf{.951}   \\
 \hline
 \end{tabular}
 }
 \caption{Comparison with the state-of-the-art methods. Results that are not available are marked with a dash.}
 \label{tab:sota}
\vspace{-2mm}
 \end{table}

\section{Conclusions}\label{sec:conclusions}
The effort to improve face recognition performance has resulted in increasingly more complex recognition pipelines. Though this process is continuously pushing performances up, the computational costs of some of these methods may not be entirely necessary. This paper turns to some of the most well established principles in image processing and computer vision -- image averaging for reduced storage and computation, and improved image quality -- seeking a simpler approach to representing sets of face images. We show that by aligning faces in 3D and partitioning them according to facial and imaging properties, average pooling provides a surprisingly effective yet computationally efficient approach to representing and matching face sets. Our system was tested on the most challenging benchmarks available today, the IJB-A and Janus CS2, demonstrating that not only does pooling compress template sizes and reduces the numbers of cross template comparisons it also lifts performances by noticeable margins. 

Our results suggest several compelling future directions. We partition faces into expert tailored bins, determined empirically to provide optimal performances. A natural question arising from this is: can these bins be optimally determined automatically? Alternatively, pooling images (rather than features extracted from them) offer opportunities for more elaborate pooling schemes. Specifically, we pooled images using a simple, non-weighted average. A potential modification would be to explore weighted averages on image {\em pixels}; that is, weighing different facial regions differently according to the information they provide. Doing so, we hope to exploit facial information from multiple, partially corrupt images.

\setlength{\parskip}{0.33em}
\section*{Acknowledgments}
This research is based upon work supported in part by the Office of the Director of National Intelligence (ODNI), Intelligence Advanced Research Projects Activity (IARPA), via IARPA 2014-14071600011. The views and conclusions contained herein are those of the authors and should not be interpreted as necessarily representing the official policies or endorsements, either expressed or implied, of ODNI, IARPA, or the U.S. Government.  The U.S. Government is authorized to reproduce and distribute reprints for Governmental purpose notwithstanding any copyright annotation thereon.

\setlength{\parskip}{0.2em}
{\small
%\bibliographystyle{ieee}
%\bibliography{pooled_faces}

}
\end{document}